\documentclass[11pt,letterpaper]{article}
\pdfoutput=1
\usepackage{naaclhlt2010}
\usepackage{color}
\usepackage{times}
\usepackage{url}
\usepackage{latexsym}
\usepackage{graphicx}
\usepackage[noend]{algorithmic}
\usepackage{algorithm}
\usepackage{rotating}
\usepackage{verbatim}
\usepackage{amsmath}
\usepackage{multirow}

\newtheorem{definition}{Definition}

\newcommand{\scrD}{\ensuremath{\mathcal{D}}}
\newcommand{\wf}{\ensuremath{\widehat{f}}}

\title{Lexical Co-occurrence, Statistical Significance, and Word Association}

\author{Dipak Chaudhari\\
  Computer Science and Engg.\\
  IIT Bombay\\
  {\small \tt dipakc@cse.iitb.ac.in}  \And
  Om P. Damani\\
  Computer Science and Engg.\\
  IIT Bombay\\
  {\small \tt  damani@cse.iitb.ac.in} \And
  Srivatsan Laxman\\
  Microsoft Research India\\
  Bangalore\\
  {\small \tt slaxman@microsoft.com}}

\date{}

\begin{document}
\maketitle

\begin{abstract}


Lexical co-occurrence is an important cue for detecting word associations. We present a theoretical
framework for discovering statistically significant lexical  co-occurrences from a given corpus. 
In contrast with the  prevalent practice of giving weightage to unigram frequencies, we focus only on the documents containing
both the terms (of a candidate bigram). We detect biases in span
distributions of associated words, while being agnostic
to variations in global unigram frequencies.
Our framework has the fidelity to distinguish different
classes of lexical co-occurrences, based on
strengths of the document and corpus-level cues of
co-occurrence in the data. We perform extensive experiments
on benchmark data sets to study the performance
of various co-occurrence measures that are
currently known in literature. We find that a relatively
obscure measure called Ochiai, and a newly introduced measure CSA capture the
notion of lexical co-occurrence best, followed next
by LLR, Dice, and TTest, while another popular
measure, PMI, suprisingly, performs poorly in the
context of lexical co-occurrence.

\end{abstract}

\section{Introduction}
\label{sec:intro}

The notion of {\em word association} is important for
numerous NLP applications, like, word sense disambiguation,
optical character recognition, speech
recognition, parsing, lexicography,
natural language generation, and machine
translation. Lexical co-occurrence is an important indicator of word association and this has
motivated several frequency-based measures for word association \cite{churchHanks89,llr,dice,cwcd}.
In this paper, we present a theoretical basis for detection and classification of lexical
co-occurrences\footnote{Note that we are interested in co-occurrence, not collocation, i.e.,
pairs of words that co-occur in a document with an arbitrary number of intervening words. Also, we use the term bigram to mean
bigram at-a-distance or spanned-bigram -- again, other words can occur in-between the constituents of
a bigram.}. In general, a lexical co-occurrence
could refer to a pair of words that occur in a large number of documents; or it could refer
to a pair of words that, although appear only in a small number of documents, occur frequently very
close to each other within each document. We formalize these ideas and construct a significance
test for co-occurrences that will allow us to detect different kinds of co-occurrences within a
single unified framework (a feature which is absent in current measures for co-occurrence). As a
by-product, our framework also leads to a better understanding of  existing measures for word
co-occurrence.



As pointed out in ~\cite{kilgariff05language}, language is never random -
which brings us to the question of what model of
random chance can give us a good statistical test
for lexical co-occurrences.
We need a null hypothesis that can account for an
observed co-occurrence as a pure chance event and
this in-turn requires a corpus generation model.
It is often reasonable to assume that documents
in the corpus are generated independent of each
other. Existing frequecy-based association
measures like PMI~\cite{churchHanks89},
LLR~\cite{llr} etc. further assume that each document
is drawn from a multinomial distribution
based on global unigram frequencies. The main
concern with such a null model is the overbearing influence of unigram
frequencies on the detection of word associations. For example, the association between {\em anomochilidae} (dwarf pipe
snakes) and {\em snake} would go undetected in our wikepedia corpus, since less than
$0.1\%$ of the pages containing {\em snake} also contained
{\em anomochilidae}. Similarly, under current models, the expected {\em span} (inter-word distance)
of a bigram is also very sensitive to the associated unigram frequencies:
the expected span of a bigram composed of low frequency
unigrams is much larger than that with
high frequency unigrams. This is contrary to
how word associations appear in language, where
semantic relationships manifest with small inter-word
distances irrespective of the underlying unigram
distributions.

These considerations motivate our search for a
more direct relationship between words,
one that can potentially be detected using careful
statistical characterization of inter-word distances, while minimizing the influence of the
associated unigram frequencies. We focus on only the documents containing both the terms (of a
candidate bigram) since in NLP applications, we often have
to chose from a set of alternatives for a given word. Hence, rather than ask the abstract
question of whether words $x$ and $y$ are related, our approach is to ask, given that $y$ is a candidate for pairing with $x$,
how likely is it that $x$ and $y$ are lexically correlated. For example, probability that  {\em
anomochilidae} is found in the vicinity of {\em snake} is higher if we knew that
{\em anomochilidae} and {\em snake} appear in the same context.


We consider a null model that represents each document as a bag of words \footnote{There can be many ways to
associate a bag of words with a document. Details of this association are not important for us,
except that the bag of words provides some kind of quantitative summary of the words within the document.}.
Then, a random permutation of
the associated bag of words gives a linear
representation for the document. An arbitrary relation between a pair
of words will result in the locations
of these words to be randomly distributed
in the documents in which they co-occur.
If the observed span distribution of a bigram resembles that under
the (random permutation) null model, then the relation between the words is not strong enough
for one word to influence the placement of the other. However, if the words are
found to occur closer together than explainable by our
null model, then we hypothesize existence of a more direct association
between these words.



In this paper, we formalize the notion of statistically significant lexical co-occurrences by introducing a
null model that can detect biases in span distributions of word associations, while being
agnostic to variations in global unigram frequencies. Our framework has the fidelity to
distinguish different classes of lexical co-occurrences, based on strengths of the document
and corpus-level cues of co-occurrence in the data.
We perform extensive  experiments on benchmark data sets to study the performance of various co-occurrence
measures that are currently known in literature. We find that a relatively obscure measure called
Ochiai, and a newly introduced measure CSA, capture the notion of lexical co-occurrence best, followed next by LLR, Dice, and TTest, while
another popular measure, PMI, suprisingly, performs poorly in the context of lexical co-occurrence.

\section{Lexically significant co-occurrences}
\label{sec:significance-test}

Consider a bigram $\alpha$. Let $\scrD=\{D_1,\ldots,D_K\}$ denote the set of 
$K$ documents (from out of the entire corpus) that contain at least one occurrence of $\alpha$.  The {\em frequency} of $\alpha$ in
document $D_i$, $f_i$, is the maximum number of {\em non-overlapped occurrences} of $\alpha$ in $D_i$. A set of occurrences of 
a bigram are called non-overlapping if the words corresponding to one occurrence from the set do not appear
in-between the words corresponding to any other occurrence from the set. 


The {\em span} of an occurrence of $\alpha$ is the `unsigned distance' 
between the first and last textual units of interest associated with that occurrence.
We mostly use words as the unit of distance, but in general, distance can be measured in
words, sentences, or even paragraphs (e.g.~an occurrence comprising two adjacent words in a sentence has a word-span of one 
and a sentence-span of zero). Likewise, the size of a document $D_i$, denoted as $\ell_i$, 
is correspondingly measured in units of words, sentences or paragraphs.
Finally, let $\wf_i$ denote the maximum number of non-overlapped occurrences of
$\alpha$ in $D_i$ with span less than a given threshold $x$. We refer to
$\wf_i$ as the {\em span-constrained frequency} of $\alpha$ in $D_i$. Note that 
$\wf_i$ cannot exceed $f_i$.


To assess the statistical significance of the bigram $\alpha$ 
we ask if the span-constrained frequency $\wf_i$ (of $\alpha$)
is more than what we would expect for it in a document of size $\ell_i$ containing $f_i$ `random' occurrences of $\alpha$.
Our intuition is that if two words are semantically related, they will often appear close to
each other in the document and so the distribution of the spans will  typically exhibit a prominent bias
toward values less than a small $x$.

Consider the null hypothesis that a document is generated as a random permutation of  the bag of words
associated with the document. Let $\pi_x(\wf,f,\ell)$ denote the probability of
observing a span-constrained frequency (for $\alpha$) of {\em at least} $\wf$ in a document of length $\ell$ that contains
a maximum of $f$ non-overlapped occurrences of $\alpha$. Observe that $\pi_x(0,f,\ell)=1$ for any
$x>0$; also, for $x\geq \ell$ we have $\pi_x(f,f,\ell)=1$ (i.e.~all $f$ occurrences will always have span
less than $x$ for $x\geq \ell$). However, for typical values of
$x$ (i.e.~for $x \ll \ell$) the probability $\pi_x(\wf,f,\ell)$ decreases with increasing $\wf$. 
For example, consider a document of length 400 with 4 non-overlapped
occurrences of $\alpha$. The probabilities of observing at least 4, 3, 2, 1 and 0 occurrences of
$\alpha$ within a span of 20 words are 0.007, 0.09, 0.41, 0.83, and 1.0 respectively. 
Since $\pi_{20}(3,4,400)=0.09$, even if 3 of the 4 occurrences of $\alpha$
(in the example document) have span less than 20 words, 
there is 9\% chance that the occurrences
were a consequence of a random event (under our null model). As a result, if
we desired a confidence-level of at least 95\%, we would have to declare $\alpha$ as {\em
insignificant}.

Given an $\epsilon$ ($0< \epsilon < 1$) and a span upper-bound $x$ ($\geq 0$) 
the document $D_i$ is said to {\em support} the hypothesis ``$\alpha$ is a $\epsilon$-significant bigram'' if $\pi_x(\wf_i,f_i,\ell) < \epsilon$. 
We refer to $\epsilon$ as the {\em document-level} lexical co-occurrence of $\alpha$.
Define indicator variables $z_i$, $i=1,\ldots,K$ as:
\begin{equation}
z_i = \left\{\begin{array}{ll}
1 & \mbox{if\ } \pi_x(\wf_i,f_i,\ell) < \epsilon \\
0 & \mbox{otherwise}\end{array} \right.
\label{eq:zi}
\end{equation}\vspace{-0.2in}

Let $Z = \sum_{i=1}^K z_i$; $Z$ models the number of documents (out of $K$) that
support the hypothesis ``$\alpha$ is a $\epsilon$-significant bigram.''
The expected value of $Z$ is given by
\begin{eqnarray}
E(Z) &=& \sum_{i=1}^K E(z_i) \label{eq:ez1}\\
	 &=& \sum_{i=1}^K \pi_x(g_\epsilon(f_i,\ell_i), f_i,\ell_i) \label{eq:ez2}
\end{eqnarray}
where $g_\epsilon(f_i,\ell_i)$ denotes the smallest $\wf$ for which we can get
$\pi_x(\wf,f_i,\ell_i)<\epsilon$ (This quantity is well-defined since $\pi_x(\wf,f_i,\ell_i)$ is
non-increasing with respect to $\wf$). For the example given earlier, $g_{0.2}(4,400)=3$
and $g_{0.05}(4,400)=4$.  
 
Using Hoeffding's Inequality, for $t>0$,
\begin{equation}
P[ Z \geq E(Z) + Kt ] \leq \exp(-2Kt^2)
\label{eq:hoeffding}
\end{equation}
Therefore, we can bound the deviation of the observed value of $Z$ from its expectation by chosing $t$ appropriately.
For example, in our corpus, the bigram ({\em canyon,\ landscape}) occurs in $K= 416$ documents. For
$\epsilon = 0.1$, we find that $Z=33$ documents (out of 416) have $\epsilon$-significant occurrences,
while $E(Z)$ is 14.34. Let $\delta = .01$. By setting $t = \sqrt{\ln{\delta}/(-2K)}=.07$, we get
$E(Z) + Kt=43.46$, which is greater than the observed value of  $Z$ (=33).
Thus, we cannot be 99\% sure that the occurrences of ({\em canyon,\ landscape}) in the 33 documents
were a consequence of non-random phenomena. Hence, our test declares  ({\em canyon,\ landscape}) as
{\em insignificant} at $\epsilon=.1, \delta=.01$. We formally state the significance test for lexical
co-occurrences next:
\begin{definition}
[Significant lexical co-occurrence] 
Consider a bigram $\alpha$ and a set of $K$ documents containing at least one occurrence of $\alpha$.
Let $Z$ denote the number of documents (out of $K$) that support the hypothesis ``$\alpha$ is
an $\epsilon$-significant bigram (for a given $\epsilon>0$, $x>0$)". 
The $K$ occurrences of the bigram $\alpha$ are regarded $\epsilon$-significant with
confidence $(1-\delta)$ (for some user-defined $\delta>0$) if we have $[Z \geq E(Z) + Kt]$, where $t=\sqrt{\log{\delta}/
(-2K)}$ and $E(Z)$ is given by Eq.~(\ref{eq:ez2}). The ratio $[Z / (E(Z) + Kt)]$ is called the
Co-occurrence Significance Ratio (CSR) for $\alpha$.
\label{def:test1}
\end{definition}


We now describe how to compute $\pi_x(\wf_i,f_i,\ell_i)$ for $\alpha$ in $D_i$. Let $N(f_i,\ell_i)$
denote the number of ways of embedding $f_i$ non-overlapped occurrences of $\alpha$ in a document of
length $\ell_i$. Similarly, let $N_x(\wf_i,f_i,\ell_i)$ denote the number of ways of embedding $f_i$
non-overlapped occurrences of $\alpha$ in a document of length $\ell_i$, in such a way that, at
least $\wf_i$ of the $f_i$ occurrences have span less than $x$. Recall that
$\pi_x(\wf_i,f_i,\ell_i)$ denotes the probability of observing a span-constrained frequency
(for $\alpha$) of at least $\wf_i$ in a document of length $\ell_i$ that contains
a maximum of $f_i$ non-overlapped occurrences of $\alpha$. Thus, we can assign the probability 
$\pi_x(\wf_i,f_i,\ell_i)$ in terms of $N(f_i,\ell_i)$ and $N_x(\wf_i,f_i,\ell_i)$ as follows:
\begin{equation}
\pi_x(\wf_i,f_i,\ell_i) = \left( \frac{N_x(\wf_i,f_i,\ell_i)}{N(f_i,\ell_i)} \right)
\label{eq:pi}
\end{equation}

To compute $N(f_i,\ell_i)$ and $N_x(\wf_i,f_i,\ell_i)$, we essentially need the histogram for $\wf$
given $f$ and $\ell$. Let $hist_{f,\ell}[\wf]$ denote the number of ways to embed $f$ non-overlapped
occurrences of a bigram in a document of length $\ell$ in such a way that exactly $\wf$ of the $f$
occurrences satisfy the span constraint $x$. We can obtain  $N(f_i,\ell_i)$ and
$N_x(\wf_i,f_i,\ell_i)$ from $hist_{f_i,\ell_i}$ using
\begin{eqnarray}
N_x(\wf_i,f_i,\ell_i) &=& \sum_{k=\wf_i}^{f_i} hist_{f_i,\ell_i}[k] \\
N(f_i,\ell_i) &=&\sum_{k=0}^{f_i} hist_{f_i,\ell_i}[k]
\end{eqnarray}

\begin{algorithm}
\small
\caption{$ComputeHist(f,\ell)$}
\label{algo-wf}
\begin{algorithmic}[1]

\REQUIRE $\ell$ - length of document; $f$ - number of non-overlapped occurrences to be embedded; 
$x$ - span constraint for occurrences

\ENSURE $hist_{f,\ell}[\cdot]$ - histogram of $\wf$ when $f$ occurrences are embedded in a document
of length $\ell$

\STATE Initialize $hist_{f,\ell}[\wf] \leftarrow 0$ for $\wf=0,\ldots,f$

\IF{$f>\ell$}
	\STATE return $hist_{f,\ell}$
\ENDIF

\IF{$f=0$}
	\STATE $hist_{f,\ell}[0] \leftarrow 1$;
	\STATE return $hist_{f,\ell}$
\ENDIF

\FOR{$i \leftarrow 1$ to $(\ell-1)$}
	\FOR{$j \leftarrow (i+1)$ to $\ell$}
		\STATE $hist_{f-1,\ell-j} \leftarrow ComputeHist(f-1, \ell-j)$
		\FOR{$k \leftarrow 0$ to $f-1$}
			\IF{$(j-i) < x$}
				\STATE $hist_{f,\ell}[k+1] \leftarrow hist_{f,\ell}[k+1] + hist_{f-1,\ell-j}[k]$
			\ELSE
				\STATE $hist_{f,\ell}[k] \leftarrow hist_{f,\ell}[k] + hist_{f-1,\ell-j}[k]$
			\ENDIF
		\ENDFOR
	\ENDFOR
\ENDFOR
\STATE return $hist_{f,\ell}$

\end{algorithmic}
\end{algorithm}

{\em Algorithm~\ref{algo-wf}} lists the pseudocode for computing the histogram $h_{f,\ell}$. It enumerates all possible ways of embedding $f$ non-overlapped
occurrences of a bigram in a document of length $\ell$. 
The main steps in the algorithm involve selecting a start and end position for
embedding the very first occurrence (lines 7-8) and then recursively calling
$ComputeHist(\cdot,\cdot)$ (line 9). The $i$-loop selects a
start position for the first occurrence of the bigram, and the $j$-loop selects the end position. The task in
the recursion step is to now compute the number of ways to embed the remaining $(f-1)$ non-overlapped occurrences in the remaining
$(\ell-j)$ positions. Once we have $hist_{f-1,\ell-j}$, we need to check whether the
occurrence introduced at positions $(i,j)$ will contribute to the $\wf$ count. If $(j-i)<x$, whenever there are $k$ span-constrained
occurrences in positions $(j+1)$ to $\ell$, there will be $(k+1)$ span-constrained occurrences in
positions 1 to $\ell$. Thus, we increment $hist_{f,\ell}[k+1]$ by the quantity
$hist_{f-1,\ell-j}[k]$ (lines 10-12). However, if $(j-i)>x$, there is no contribution
to the span-constrained frequency from the $(i,j)$ occurrence, and so we increment $hist_{f,\ell}[k]$
by the quantity $hist_{f-1,\ell-j}[k]$ (lines 10-11, 13-14).

This algorithm is exponential in $f$ and $l$, but it
does not depend explicitly on the data. This allows us to populate the histogram off-line, and
publish the $\pi_x(\wf,f,\ell)$ tables for various $x$, $\wf$, $f$ and $\ell$. 
(If the paper is accepted, we will make an interface to this table publicly available).

\section{Utility of CSR test}\label{sec:discussion}


Evidence for significant lexical co-occurrences can be gathered at two levels in the data --
document-level and corpus-level. First, at the document
level, we may find that a surprisingly high proportion of occurrences {\em within} a
document (of a pair of words)  have smaller spans than they would by random chance. Second, at the corpus-level, we may find
a pair of words appearing closer-than-random in an unusually high number of documents in the
corpus. The significance test of {\em Definition~\ref{def:test1}} is capable of gathering both kinds
of evidence from data in carefully calibrated amounts. Prescribing $\epsilon$ essentially fixes the strength of the document-level
hypothesis in our test. A small $\epsilon$ corresponds to a strong document-level hypothesis and
vice-versa. The second parameter in our test, $\delta$, controls the confidence of our decision
given all the documents in the data corpus. A small $\delta$
represents a high confidence test (in the sense that there are a surprisingly large number of documents in
the corpus, each of which, individually have some evidence of relatedness for the pair of words). 
%
By running the significance test with different values of $\epsilon$ and $\delta$, we can detect
different types of lexically significant co-occurrences. We illustrate the utility of
our test of significance by considering the 4 types of lexical significant co-occurrences

 {\em Type A}: These correspond to the strongest lexical co-occurrences in the data, with strong
document-level hypotheses (low $\epsilon$) as well as high corpus-level confidence (low $\delta$). Intuitively, if a pair of
words appear close together several times within a document, and if this pattern is observed in a large
number of documents, then the co-occurrence is of {\em Type A}. 


{\em Type B}: These are co-occurrences based on weak document-level hypotheses (high $\epsilon$) 
but because of repeated observation in a substantial number of documents in the corpus, we can still detect them with
high confidence (low $\delta$). We expect many interesting
lexical co-occurrences in text corpora to be of
Type B – pairs of words that appear close to each
other only a small number of times within a document,
but they appear together in a large number of documents.

{\em Type C}: Sometimes we may be interested in words that are
strongly correlated within a document, even if we observe the strong correlation only in a
relatively small number of documents in the corpus. These correspond to Type C co-occurrences. 
Although they are statistically weaker inferences than
those of Type A and Type B (since confidence $(1-\delta)$ is lower) Type C co-occurrences represent an important class of relationships
between words. If the document corpus contains a very small of number documents on some topic, then
strong co-occurrences (i.e. those found with low $\epsilon$)  which are unique to that topic may not be
detected at low values of $\delta$. By relaxing the confidence parameter $\delta$, we may be able to detect
such occurrences (possibly at the cost of some extra false positives). 


{\em Type D}: These co-occurrences represent the weakest correlations found in
the data, since they neither employ a strong document-level hypothesis nor enforce a high 
corpus-level confidence. In most applications, we expect Type D co-occurrences to be of little use, with their best case
utility being to provide a baseline for disambiguating Type C co-occurrences. 


\begin{table}
\centering
\begin{tabular}{|l|l|l|}   
 \hline
Type	&	$\epsilon$	&	$\delta$ \\ \hline
A		&	$\leq 0.1$	& $\leq 0.1$ \\
B		&	$\geq 0.4$	& $\leq 0.1$ \\
C		&	$\leq 0.1$	& $\geq 0.4$ \\
D		&	$\geq 0.4$	& $\geq 0.4$ \\ \hline
\end{tabular}
\caption{4 types of lexical co-occurrences.}
\label{tab:edpairs}
\end{table}

In the experiments we describe later, we fix the $\epsilon$ and $\delta$ for the different Types as
per Table~\ref{tab:edpairs}. Finally, we note that Types B and C subsume Type A; similarly, Type D
subsumes all three other types. Thus, to detect co-occurrences that are exclusively of (say) Type B,
we would have to run the test with a high $\epsilon$ and low $\delta$ and then remove from the
output, those co-occurrences that are also part of Type A.

\section{Experimental Results}

\subsection{Datasets and Text Corpus}

Since similarity and relatedness are
different kinds of word associations \cite{budanitskyHirst}, in ~\cite{simRelDatasets}
two different data sets, namely 203 words {\em sim} (the
union of similar and unrelated pairs) and 252 words {\em rel} (the union of related
and unrelated pairs) datasets are derived from {\em wordsim}~\cite{wordsim353}.
We use these two data sets in our experiments. These datasets are symmetric in that the order of words in a pair is not expected to matter. As some of our chosen co-occurrence measures are asymmetric, we also report results on the asymmetric 272-words {\em esslli} dataset
for the `free association' task at~\cite{esslli08}. 


We use the Wikipedia~\cite{wikipedia} corpus in our experiments. It contains 2.7 million articles
for a total size of 1.24 Gigawords.  We did not pre-process the corpus - no lemmatization,
no function-word removal. When counting document size in words, punctuation symbols were ignored.
Documents larger than 1500 words were partitioned keeping the size of each part to no greater
than 1500 words.

In Table~\ref{tab:typeExamples}, we present some examples of
different types of co-occurrences observed in the data. 


\begin{table*}
\centering
{\scriptsize
\begin{tabular}{|l|l|l|l|l|}
\hline
Dataset & Type A bigrams & Type B bigrams & Type C bigrams & Type D bigrams \\ \hline
\multirow{2}{*}{sim} & announcement-news & forest-graveyard & lobster-wine & stock-egg \\
 & bread-butter & tiger-carnivore & lad-brother & cup-object \\ \hline
\multirow{2}{*}{rel} & baby-mother & alcohol-chemistry & victim-emergency & money-withdrawal \\
 & country-citizen & physics-proton & territory-kilometer & minority-peace \\ \hline
\multirow{2}{*}{esslli} & arrest-police & pamphlet-read & meditate-think & fairground-roundabout \\
 & arson-fire & spindly-thin & ramble-walk & \\
\hline
\end{tabular}
\label{tab:typeExamples}
 \caption{Examples of Type A, B, C and D co-occurrences under a span constraint of 20 words.}
}
\end{table*}

\subsection{Performance of different co-occurrence measures}

We now compare the performance of various frequency-based measures in the context of lexical
significance. Given the large numbers of measures proposed in the literature~\cite{pecina06}, we need to identify a subset of measures to compare.
Inspired by \cite{ecologyMeasures} and \cite{dataMiningTan} we identify three properties
of co-occurrence measure which may be useful for language processing applications. First is {\em Symmetry} - does the measure yield the same association score for (x,y) and (y,x)? Second is {\em Null Addition} - does addition of data containing neither x nor y affect the association
score for (x,y)? And, finally, {\em Homogenity} - if we replicate the corpus several times and merge them to construct a
larger corpus, does the association score for (x,y) remain unchanged? Note that the concept of homogenity conflicts
with the notion of statistical support, as support increases in direct proportion with the absolute amount of evidence.
Different applications may need co-occurrence measures having different combinations of these properties. 

\begin{table}
\scriptsize
\begin{tabular}{|p{1.8cm}| p{2.95cm}|l|l|l|}   
 \hline
Method & Formula & \begin{sideways}Symm. \end{sideways} & \begin{sideways}Null Add. \end{sideways}& \begin{sideways}Homo. \end{sideways}\\
\hline
CSR (this work) & $Z / (E(Z) + Kt)$ & Y & Y &  N\\ \hline 
CSA (this work) &  $\frac{\hat{f}(x,y)}{\sqrt{K}} $  & Y & N &  Y\\ \hline
LLR~\cite{llr} & ${\displaystyle{\sum_{x', y' }}}p(x',y')log\frac{p(x',y')}{p(x')p(y')}$ & Y & Y  & Y \\ \hline
PMI~\cite{churchHanks89} & $log\frac{p(x,y)}{p(x)p(y)}$ & Y & N &  Y \\ \hline
SCI~\cite{cwcd} & $\frac{p(x,y)}{p(x)\sqrt{p(y)}}$ & N & N  &  Y\\ \hline
CWCD~\cite{cwcd} & $\frac{\hat{f}(x,y)}{p(x)}\frac{1/max\left(p(x),p(y)\right)}{M}$ & N & N  &  Y\\ \hline
Pearson's $\chi^2$ test & ${\displaystyle\sum_{x',y'}} \frac{\left(\hat{f}(x',y')-E\hat{f}(x',y')\right)^2}{E\hat{f}(x',y')}$ & Y & Y &  Y\\ \hline
T-test & $\frac{\hat{f}(x,y)-E\hat{f}(x,y)}{\sqrt{\hat{f}(x,y)\left(1-\frac{\hat{f}(x,y)}{N}\right)}}$ & Y & N & Y\\ \hline
Dice~\cite{dice} & $\frac{2\hat{f}(x,y)}{f(x)+f(y)}$ & Y & N &  Y\\ \hline
Ochiai~\cite{ecologyMeasures} & $\frac{\hat{f}(x,y)}{\sqrt{f(x)f(y)}}$ & Y & N  & Y\\ \hline
Jaccard~\cite{jaccard} & $\frac{\hat{f}(x,y)}{f(x)+f(y)-\hat{f}(x,y)}$ & Y & N  & Y \\ \hline
\end{tabular}
{\scriptsize
Terminology: ($x' \in \{x,\neg x\}$ and
$y' \in\{y,\neg y\}$) \\
\begin{tabular}{l l }
$N$ & Total number of tokens in the corpus \\
$f(x),f(y)$ & unigram frequencies of $x,y$ in the corpus \\
$p(x),p(y)$ & $f(x)/N,f(y)/N $\\
$\hat{f}(x,y)$ & Span-constrained ($x,y$) bigram frequency\\
$\hat{p}(x,y)$ & $\hat{f}(x,y)/N $\\
$M$ & Harmonic mean of the spans of $\hat{f}(x,y)$ occurrences\\
$E\hat{f}(x,y)$ & Expected value of f(x,y) \\
\hline
\end{tabular}
}
\caption{ \small Properties of selected co-occurrence measures }
\label{tab:methods}
\end{table}


Table~\ref{tab:methods} shows the characteristics of our chosen co-occurrence measures, which were selected from several domains like ecology,
psychology, medicine, and language processing. Except Ochiai~\cite{Ochiai}, \cite{ecologyMeasures}, and the recently introduced measure CWCD~\cite{cwcd}\footnote{From various so-called windowless measures introduced in~\cite{cwcd}, we chose the best-performing variant Cue-Weighted Co-Dispersion (CWCD) and implemented a window based version of it with harmonic mean. We note that any of windowless (or spanless) measure can easily be thought of as a special case of a window-based measure where the windowless formulation corresponds to a very large window (or span in our terminology).}, all other selected measures are well-known in the NLP community~\cite{pecina06}.
 Based on our extensive study of theoretical and empirical properties of CSR, we also introduce a new bigram frequency based measure called CSA ({\em Co-occurrence Significance Approximated}), which approximates the behaviour of CSR over a wide range of parameter settings.

In our experiments, we found that Ochiai and Chi-Square have almost identical performance, differing only in 3rd decimal digits.
This can be be explained easily. In our context, for any word $x$,  as defined in Table~\ref{tab:methods},
$f(x) << N$ and therefore $p(x) << 1$. With this, Chi-Square reduces to square of Ochiai. Similarly Jaccard and Dice coincide,
since $f(x,y) << f(x)$ and $f(x,y) << f(y)$. Hence we do not
report further results for Chi-Square and Jaccard.

In our first set of experiments, we compared the performance of various frequency-based
measures in terms of their suitability for detecting lexically significant co-occurrences 
(cf.~{\em Definition~\ref{def:test1}}). 
A high Spearman correlation coefficient between the ranked list produced by a given measure and the list produced by CSR with respect to some choice of
$\epsilon$ and $\delta$ would imply that the measure is effective in detecting the corresponding {\em type} of
lexically significant co-occurrences.

\begin{table}
\centering
\scriptsize
\begin{tabular}{|l|l|l|l|l|}
\hline
						& 			& \multicolumn{3}{|c|}{Span Threshold} \\ \hline
	Measure				&	Data	&	 5w		&	25w		&	50w	\\ \hline
\multirow{3}{*}{PMI}	& sim		& C			& -			& - \\
						& rel		& -			& -			& -	\\
						& essli		& -			& -			& -	\\ \hline
\multirow{3}{*}{CWCD}	& sim		& -			& -			& - \\
						& rel		& -			& -			& -	\\
						& essli		& -			& -			& -	\\ \hline
\multirow{3}{*}{CSA}	& sim		&A, B, C, D	& A, B, C& A, B, C \\
						& rel		&A, B, C, D	& A, B, C	& A, C	\\
						& essli		&A, B, C, D	& A, B, C		& A, C	\\ \hline
\multirow{3}{*}{Dice}	& sim		&A, B, C, D	& A, B, C	& A, B \\
						& rel		&A, B, C, D	& -			& -	\\
						& essli		& -			& -			& -	\\ \hline
\multirow{3}{*}{Ochiai}	& sim		&A, B, C, D	& A, B, C, D& A, B, C \\
						& rel		&A, B, C, D	& A, B, C	& A, B, C	\\
						& essli		&A, B, C, D	& A, B		& A	\\ \hline
\multirow{3}{*}{LLR}	& sim		&A, B, C, D	& A, B		& A \\
						& rel		&A, B, C, D	& A			& A	\\
						& essli		&A, B, C	& A			& A	\\ \hline
\multirow{3}{*}{TTest}	& sim		&A, B, C	& A			& - \\
						& rel		&A, B, C	& -			& -	\\
						& essli		& -			& -			& -	\\ \hline
\multirow{3}{*}{SCI}	& sim		& -			& -			& - \\
						& rel		& -			& -			& -	\\
						& essli		& -			& -			& -	\\ \hline
\end{tabular}
\caption{Types of lexical co-occurrences detected by different measures}
\label{tab:ABCDsummary}
\end{table}

\begin{figure*}
  \begin{center}
\resizebox{160mm}{!}
{\includegraphics{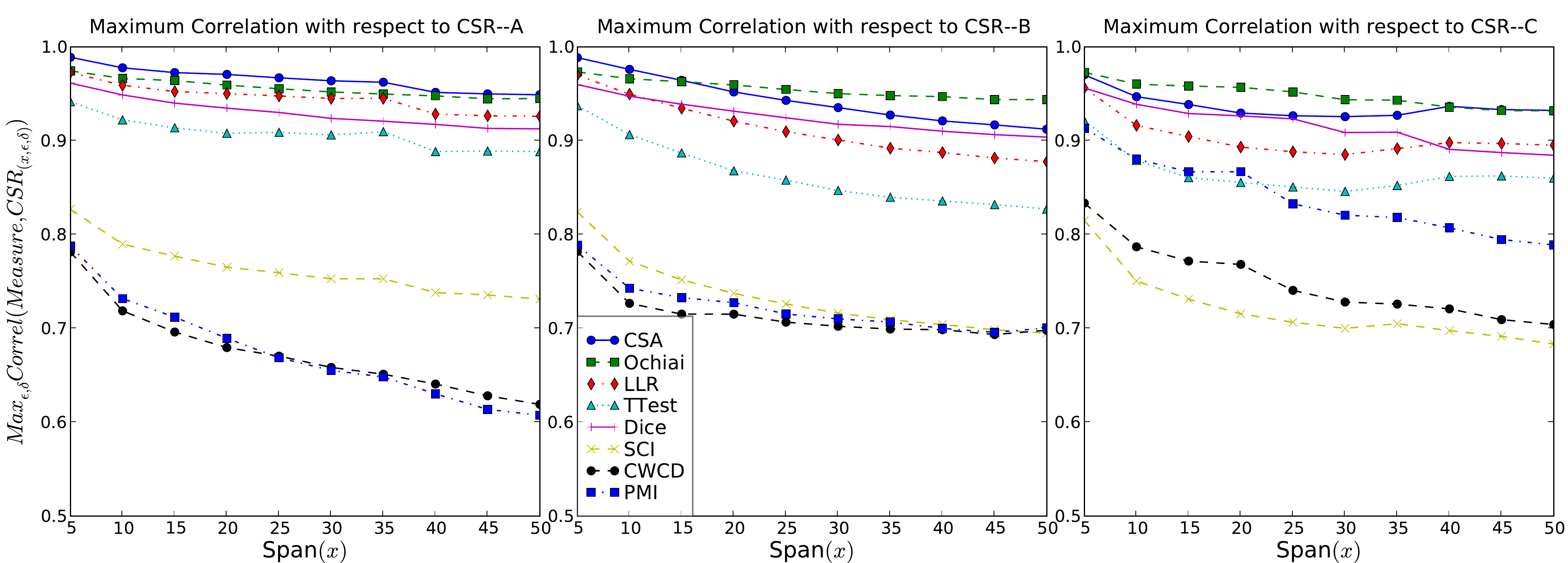}} \\
    \caption{\small Maximum correlation of various measures with various types of CSR for sim dataset}
    \label{fig:maxCorSim}
  \end{center}
\end{figure*}

The Table~\ref{tab:ABCDsummary} lists
for each measure and for each data set, the different types of lexically significant co-occurrences that the
measure is able to detect effectively -- if the corresponding Spearman
correlation coefficient exceeds 0.90, we consider the measure to be effective for the given
type. Results are shown for three different span constraints --
small span of 5 words (or 5w), medium span of 25 words (or 25w) and large span of 50 words (or 50w).
For example, the CSA and Ochiai measures are effective in detecting all 4 types of lexically significant
co-occurrences (A, B, C and D) in all three data sets, when the span constraint is set to 5 words.
Figure~\ref{fig:maxCorSim} presents a detailed quantitative comparison of the best performance of each
measure with respect to each type of co-occurrence for a range of different span constraints on the
sim data set (Similar results were obtained on other data sets). The inferences we can draw are
consistent with the results of Table~\ref{tab:ABCDsummary}.

\begin{table}
\scriptsize
\centering

\begin{tabular}{|l|l|l|l|l|l|}
\hline
						& 			& \multicolumn{4}{|c|}{Parameters for best correlation} \\ \hline
	Measure				&	Span	&	$\epsilon$	&	$\delta$	& Type	& Correlation		\\ \hline
\multirow{3}{*}{PMI}	& 5w		& 0.05		& 1			& C		& 91.3 \\
						& 25w		& 0.40		& 1			& D		& 85.3	\\
						& 50w		& 0.50		& 1			& D		& 82.0 \\ \hline
\multirow{3}{*}{CWCD}	& 5w		& 0.99		& 0.9		& D 	& 83.6 \\
						& 25w		& 0.50		& 0.9		& D		& 76.0	\\
						& 50w		& 0.50		& 0.9		& D		& 74.4	\\ \hline
\multirow{3}{*}{CSA}	& 5w		& 0.1		& 0.0005		& A 	& 98.9 \\
						& 25w		& 0.05		& 0.0005		& A		& 96.7	\\
						& 50w		& 0.1		& 0.0005		& A		& 94.9	\\ \hline
\multirow{3}{*}{Dice}	& 5w		& 0.1		& 0.005		& A		& 96.1 \\
						& 25w		& 0.05		& 0.005		& A		& 93.0	\\
						& 50w		& 0.1		& 0.0005	& A		& 91.3 \\ \hline
\multirow{3}{*}{Ochiai}	& 5w		& 0.1		& 0.1		& A		& 97.4 \\
						& 25w		& 0.1		& 0.01		& A		& 95.5	\\
						& 50w		& 0.1		& 0.005		& A		& 94.5 \\ \hline
\multirow{3}{*}{LLR}	& 5w		& 0.05		& 0.0005	& A 	& 97.3 \\
						& 25w		& 0.05		& 0.0005	& A		& 94.8 \\
						& 50w		& 0.1		& 0.0005	& A		& 92.6 \\ \hline
\multirow{3}{*}{TTest}	& 5w		& 0.05		& 0.0005	& A 	& 94.2 \\
						& 25w		& 0.05		& 0.0005	& A		& 90.9 \\
						& 50w		& 0.1		& 0.0005	& A		& 88.8 \\ \hline
\multirow{3}{*}{SCI}	& 5w		& 0.05		& 0.0005	& A 	& 82.7 \\
						& 25w		& 0.05		& 0.0005	& A		& 75.9 \\
						& 50w		& 0.1		& 0.0005	& A		& 73.1 \\ \hline
\end{tabular}
\caption{Best performing $(\epsilon,\delta)$-pairs for different measures on {\em sim} data}
\label{tab:topsummarysim}
\end{table}

In our next experiment, we examine which of the four types of co-occurrences are best captured by each measure.
Results for the sim data set are listed in Table~\ref{tab:topsummarysim} (Similar results were obtained on the other data sets). For each
measure and for each span constraint, the table describes the best performing parameters ($\epsilon$
and $\delta$), the corresponding co-occurrence Type and the associated `best' correlation achieved with respect to
the test of {\em Definition~\ref{def:test1}} .
The results show that, irrespective of the span
constraint, most measures perform best on Type A co-occurrences. This is reasonable because
Type A essentially represents the strongest correlations in the data and one would expect the
measures to capture the strong correlations better than weaker ones. There are however, two
exceptions to this rule, namely PMI and CWCD, which instead peak at Types C or D. The best correlations for
these two measures are also typically lower than the other measures. We now summarize the main
findings from our study:
\begin{itemize}

\item The relatively obscure Ochiai, and the newly introduce CSA are the best performing measure, in terms of detecting all types
of lexical co-occurrences in all data sets and for a wide range of span constraints.

\item Dice, LLR and TTest are the other measures that effectively track lexically significant
co-occurrences (although, all three are less effective as the span constraints become larger).

\item SCI, CWCD, and the popular PMI measure
are ineffective at capturing {\em any} notion of lexically significant co-occurrences, even for small
span constraints. In fact, the best result for PMI is the detection of Type C co-occurrences in the
sim data set. The low $\epsilon$ and high $\delta$ setting of Type C suggests that PMI does a poor
job of detecting the strongest co-occurrences in the data, overlooking both strong document-level as
well as corpus-level cues for lexical significance. 


\end{itemize}

\begin{table*}
\begin{center}
 \scriptsize \addtolength{\tabcolsep}{-5pt}
	\begin{tabular}{|c | c || c | c |l| c | c || c | c |l| c | c || c | c |}  \cline{1-4} \cline{6-9} \cline{11-14}
  \multicolumn{4}{|c|}{sim} && \multicolumn{4}{|c|}{rel} && \multicolumn{4}{|c|}{esslli} \\ \cline{1-4} \cline{6-9} \cline{11-14}
PMI top 10 & R & Ochiai top 10 & R && PMI top 10 & R & Ochiai top 10 & R && PMI top 10 & R & Ochiai top 10 & R\\ \cline{1-4} \cline{6-9} \cline{11-14}
vodka-gin & 42 & football-soccer & 3 &  & money-laundering & 2 & soap-opera & 1 &  & nook-cranny & 91 & floyd-pink & 4 \\
seafood-lobster & 59 & street-avenue & 5 &  & soap-opera & 1 & money-laundering & 2 &  & hither-thither & 104 & either-or & 1 \\ 
bread-butter & 13 & physics-chemistry & 2 &  & opec-oil & 8 & computer-software & 18 &  & sprain-ankle & 60 & election-general & 7 \\ 
vodka-brandy & 99 & television-radio & 6 &  & weather-forecast & 5 & television-film & 7 &  & blimey-cor & 147 & nook-cranny & 91 \\ 
midday-noon & 79 & championship-tournament & 10 &  & psychology-cognition & 77 & jerusalem-israel & 16 &  & margarine-butter & 77 & twentieth-century & 2 \\ 
murder-manslaughter & 19 & man-woman & 16 &  & decoration-valor & 73 & weather-forecast & 5 &  & tinker-tailor & 65 & bride-groom & 16 \\ 
cucumber-potato & 130 & vodka-gin & 42 &  & gender-equality & 11 & drug-abuse & 4 &  & ding-dong & 26 & you-me & 14 \\ 
dividend-payment & 61 & king-queen & 9 &  & tennis-racket & 20 & credit-card & 3 &  & bride-groom & 16 & north-south & 19 \\ 
physics-chemistry & 2 & car-automobile & 43 &  & liability-insurance & 25 & game-series & 12 &  & jigsaw-puzzle & 30 & question-answer & 11 \\ 
psychology-psychiatry & 27 & harvard-yale & 11 &  & fbi-investigation & 10 & stock-market & 9 &  & bidder-auction & 76 & atlantic-ocean & 10 \\ \cline{1-4} \cline{6-9} \cline{11-14}
\end{tabular}
\caption{Top 10 bigrams according to PMI and Ochiai rankings on \emph{sim}, \emph{rel}, and \emph{esslli} datasets. 'R' denotes the bigrams rankings according to type-A CSR measure($\epsilon=0.1, \delta=0.1$). Span of 25 words is used for all the three measures. }
\label{tab:top10PMIOchiai}
\end{center}
\end{table*}

Note that our results do not contradict the utility of PMI, SCI, or, CWCD as word-association
measures. We only observe their poor performance in context of detecting lexical co-occurrences. Also, our notion of lexical co-occurrence is symmetric.
It is possible that asymmetric SCI may have competitive performance for certain asymmetric tasks compared to the better performing symmetric measures.
Finally, to give a qualitative feel about the differences in the correlations preferred by different methods, 
in Table~\ref{tab:top10PMIOchiai}, we show the top 10 bigrams picked by PMI and Ochiai for all three datasets.

\section{Relation between lexical co-occurrence and human judgements}

\begin{table*}
\scriptsize \addtolength{\tabcolsep}{-5pt}
\begin{tabular}{|l|c|c|c|c|c|c|c|c|c|c|}
\hline
Method & 1 & 2 & 3 & 4 & 5 & 6 & 7 & 8 & 9 & 10\tabularnewline
\hline
Human & environment & maradona & opec & computer & money & jerusalem &
law & weather & network & fbi\tabularnewline
Judgement & ecology (84)& football (53)& oil (8)& software (18)& bank (28)& israel (16)&
lawyer (42)& forecast (5)& hardware (107)& investigation (10)\tabularnewline
\hline
\multirow{2}{*}{CSR} & soap & money  & credit & drug  & weather & cup & television &
opec & stock & fbi\tabularnewline
 & opera (24) & laundering (129)& card (20)& abuse (69)& forecast (8)& coffee (82)& film (31)& oil (3)
& market (19)& investigation (10)\tabularnewline
\hline
\end{tabular}
\caption{ Top 10 word associations picked in rel dataset. The numbers in the brackets are the cross rankings: CSR rankings in the human row and human rankings in the CSR row. CSR parameters are same as that for Table~\ref{tab:top10PMIOchiai}. }
\label{top10humanCSR}
\end{table*}

While the focus of our work is on characterizing the statistically significant lexical co-occurrence, as illustrated in in Table~\ref{top10humanCSR}, human judgement of word association is governed by many factors in addition to lexical co-occurrence considerations, and many non co-occurrence based measures have been designed to capture semantic word association. Notable among them are distributional similarity based measures \cite{simRelDatasets,bollegalaMI07,chenLW06}
and knowledge-based measures \cite{wikiLinkMeasure,hughes_lexical_2007,Gabrilovich07computingsemantic,wikiwalk09,wikirelate,wordsim353,lsaEsslli08}. Since our focus is on frequency based measures alone, we do not discuss these other measures.

The lexical co-occurrence phenomenon and the human judgement of semantic association are related but different dimensions of relationships between words and different applications may
prefer one over the other. For example, suppose, given one word (say {\em dirt}), the task is to choose from among a number of
alternatives for the second(say {\em grime} and {\em filth}). Human judgment scores for {\em (dirt, grime)} and {\em (dirt, filth)}
are 5.4 and 6.1  respectively. However, their lexical co-occurrence scores (CSR) are 1.49 and 0.84 respectively. This is because {\em filth} is often used in a moral context as well.  {\em Grime} is usually used
only in a physical sense.  {\em Dirt} is used mostly in a physical sense,
but is a bit more generic and may be used in a moral sense
occasionally. Hence {\em (dirt, grime)} is more correlated in corpus than {\em (dirt, filth)}. This shows that human judgement is fallible and annotators may ignore the subtleties of meanings that may be picked up by a statistical techniques like ours.

In general, for association with a given word, all synonyms of a second word will be given similar semantic
relatedness score by human judges but they may have very different lexical association scores. 

For  applications where the notion of statistical lexical
co-occurrence is potentially more relevant than semantic relatedness, our method can be used to generate a gold-standard of lexical association (against which other association measures can be evaluated). In this context, it is interesting to note that contrary to the human judgement, each one of the co-occurrence measures studied by us finds {\em (dirt, grime)} more associated than {\em (dirt, filth)}.


Having explained that significant lexical co-occurrence is a fundamentally different notion than human judgement of word association, we also want to emphasize that the two are not completely different notions either and they correlate reasonably well with each-other.
For {\em sim, rel}, and {\em essli} datasets, CSR's best correlations with human judgment are 0.74, 0.65, and 0.46 respectively. Note that CSR is a symmetric notion and hence correlates far more with human judgement for symmetric {\em sim} and {\em rel} datasets than for the asymmetric {\em essli} dataset.
Also, at first glance, it is little counter-intuitive that the notion
of lexical co-occurrence yields better correlations with the sim (based on {\em similarity}) data set when compared to
the rel(based on {\em relatedness}) data set. This can essentially be explained by our observation that
similar words tend to co-occur less frequently by-chance than the related words.

\section{Conclusions}

In this paper, we introduced the notion of statistically significant lexical co-occurrences. We
detected skews in span distributions of bigrams to assess significance and showed how our method
allows classification of co-occurrences into different types. We performed experiments to assess the
performance of various frequency-based measures for detecting lexically signficant co-occurrences.
We believe lexical co-occurrence can play a critical role in several applications, including sense disambiguation, mutli-word spotting, etc. We will address some of these in our future work.

\bibliographystyle{coling}
\bibliography{references}

\end{document}